\documentclass[10pt,twocolumn,letterpaper]{article}

\usepackage{cvpr}       

\usepackage{graphicx}
\usepackage{amsmath}
\usepackage{amssymb}
\usepackage{bbm}
\usepackage{booktabs}
\usepackage{multirow}
\usepackage[table,xcdraw]{xcolor}
\usepackage{float}
\usepackage{subcaption}

\usepackage[pagebackref,breaklinks,colorlinks]{hyperref}
\usepackage{paralist}

\usepackage[capitalize]{cleveref}
\crefname{section}{Sec.}{Secs.}
\Crefname{section}{Section}{Sections}
\Crefname{table}{Table}{Tables}
\crefname{table}{Tab.}{Tabs.}

\def\cvprPaperID{6684} 
\def\confName{CVPR}
\def\confYear{2022}

\begin{document}

\title{Towards End-to-End Unified Scene Text Detection and Layout Analysis}

\author{Shangbang Long, Siyang Qin, Dmitry Panteleev, Alessandro Bissacco, Yasuhisa Fujii, Michalis Raptis \\
Google Research\\
{\tt\small \{longshangbang,qinb,dpantele,bissacco,yasuhisaf,mraptis\}@google.com}

}
\maketitle

\begin{abstract}
Scene text detection and document layout analysis have long been treated as two separate tasks in different image domains.
In this paper, we bring them together and introduce the task of unified scene text detection and layout analysis. The first hierarchical scene text dataset is introduced to enable this novel research task.
We also propose a novel method that is able to simultaneously detect scene text and form text clusters in a unified way. 
Comprehensive experiments show that our unified model achieves better performance than multiple well-designed baseline methods. 
Additionally, this model achieves state-of-the-art results on multiple scene text detection datasets without the need of complex post-processing. 
Dataset and code: \url{https://github.com/google-research-datasets/hiertext}.
\end{abstract}

%%%%%%%%% BODY TEXT

% ----- ----- ----- ----- ----- Introduction ----- ----- ----- ----- ----- 
\section{Introduction}
\label{sec:intro}
The ability to read and understand text in natural scenes and digital documents plays an important role in anthropocentric applications of computer vision. 
While state-of-the-art text detection systems such as \cite{sheng2021centripetaltext,Zhang_2021_ICCV} excel at localizing individual text entities, visual text understanding \cite{appalaraju2021docformer} requires comprehension of the semantic and geometric layout \cite{cattoni1998geometric,breuel2002two} of the textual content. 
In the current literature, most works focus on the individual tasks of text entities detection \cite{Zhang_2021_ICCV,he2021most,baek2019character} and layout analysis \cite{yang2017learning,lee2019page} in a separate way, devoting
all the power of deep learning models to task-specific performance. 
We argue that joint treatment of these two closely related problems can result not only in simpler and more efficient models, but also models that are more accurate across all tasks.
Additionally, an all-in-one, unified text and layout detection architecture can become indispensable for text reasoning tasks such as text-based VQA \cite{singh2019towards, biten2019scene} and image captioning \cite{xu2021towards}.

\begin{figure}[t]
  \centering
  \includegraphics[width=1.0\linewidth]{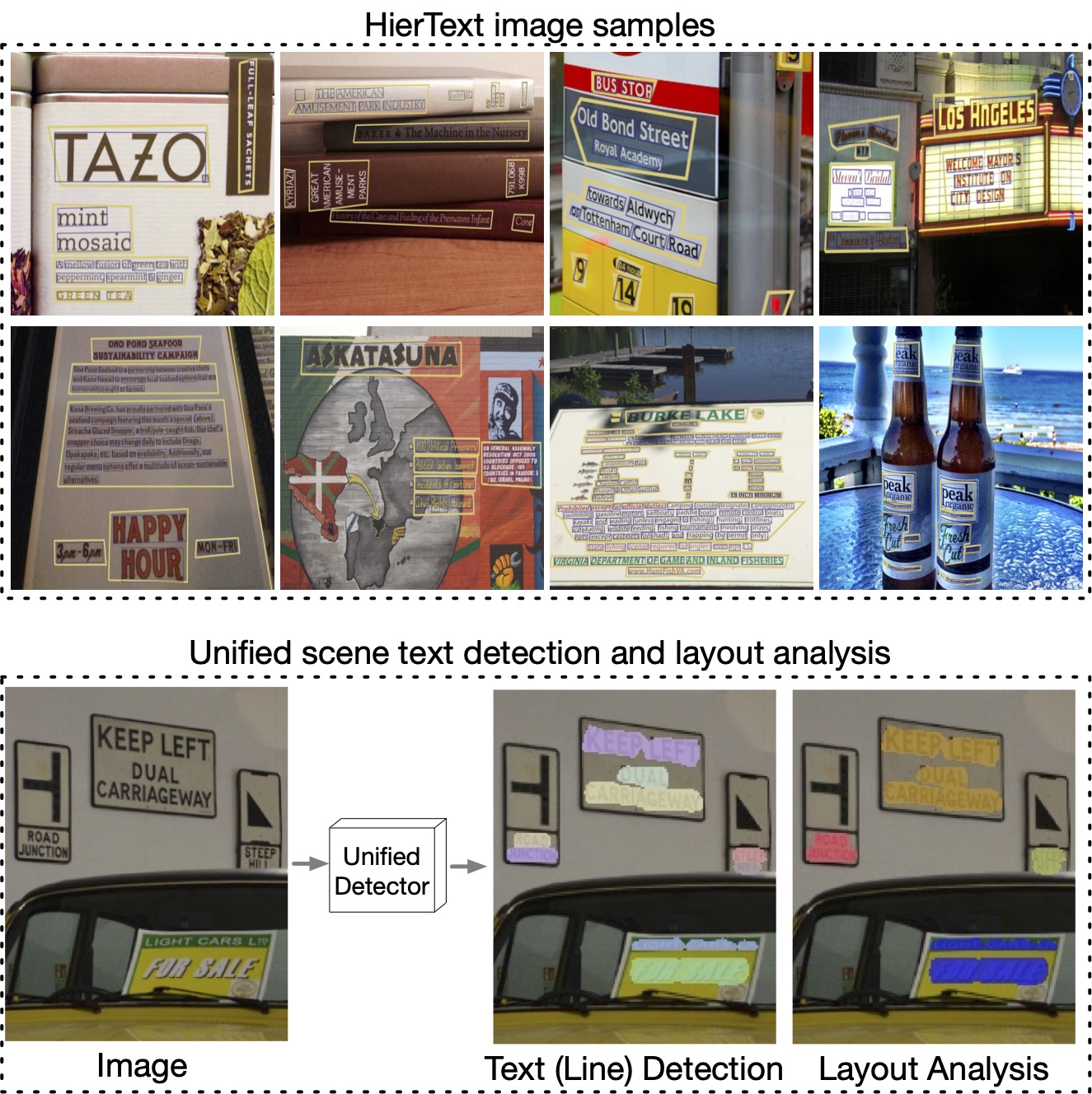}
  \caption{ 
  \textbf{Top}: We introduce the task of unified text detection and layout analysis, and collect a dataset called \textbf{HierText} with hierarchical annotations. {\color{blue}Blue} boxes are word level bounding boxes. {\color{yellow}Yellow} boundaries mark the ground-truth clustering of text. 
  Line-level annotations and transcriptions are not visualized to avoid overcrowding.
  \textbf{Bottom}: We propose an end-to-end model called \textbf{Unified Detector} which can simultaneously detect text as masks and further group them into clusters. The model produces masks for text detection and an affinity matrix to cluster text lines, both in an end-to-end fashion without complex post-processing.
  We visualize the layout analysis results by coloring text line masks according to their clusters.
  }
  \label{fig:dataset}
   \vspace{-2.5mm}
\end{figure}

The division between text detection and geometric layout analysis tasks has led to parallel and separate research directions. 
Text detectors \cite{Dai_2021_CVPR,he2021most,Zhang_2021_ICCV,qin2019towards} usually treat word-level annotations, i.e. sequence of characters not interrupted by space, as the only supervision signal. Conversely, geometric layout analysis algorithms \cite{appalaraju2021docformer,zhong2019publaynet,wang2022post,lee2019page,yang2017learning} focus on digital documents and either assume word-level text information as given \cite{appalaraju2021docformer,yang2017learning,wang2022post} or directly predict geometric structures without reasoning for their atomic elements \cite{zhong2019publaynet}. 
We ask: \emph{Can there be a reconciliation of text entity detection and geometric layout analysis? 
Can geometric layout analysis target both natural scenes and digital documents?} 
These questions are important given their relevance in real-world applications, such as screen readers for visually impaired and image-based translation.

Our work aims to unify text detection and geometric layout analysis. 
We introduce a new image dataset called \textbf{HierText}. 
It is the first dataset featuring hierarchical annotations of text in natural scenes and documents (Fig.~\ref{fig:dataset}, top).
The dataset contains high quality \textit{word}, \textit{line}, and \textit{paragraph} level annotations. ``Text lines'' are defined as logically connected sequences of words that are aligned in spatial proximity. 
Text lines that belong to the same semantic topic and are geometrically coherent form ``paragraphs''. 
Images in HierText average more than $100$ words per image, twice denser than the current highest density scene text dataset \cite{Singh_2021_CVPR}.
Experimental results show our dataset is complementary to other public datasets \cite{karatzas2015icdar, ch2017total,yuliang2017detecting,yao2012detecting,nayef2017icdar2017,nayef2019icdar2019,sun2019icdar,chng2019icdar2019,Singh_2021_CVPR,krylov2021open} for the standalone text detection task.

In addition to HierText, we present a novel model, \textbf{Unified Detector}, that simultaneously detects text entities and performs layout analysis by grouping text entities, as illustrated in the bottom of Fig.~\ref{fig:dataset}. 
Unified detector consolidates an end-to-end instance segmentation model, MaX-DeepLab \cite{wang2021max}, to detect arbitrarily shaped text and multi-head self-attention layers\cite{vaswani2017attention} to form text clusters.
The proposed model enables end-to-end training and inference with a single-stage simplified pipeline. 
It eliminates complex label generation processes \cite{baek2019character,sheng2021centripetaltext} during training and sophisticated post-processing\cite{long2018textsnake,zhou2017east} during inference.  
Unified Detector outperforms competitive baselines and even a commercial solution on the task of unified text detection and geometric layout analysis, demonstrating its effectiveness. 

Along with the unified task, we also evaluate our model on the standalone scene text detection task using existing public datasets, including ICDAR 2017 MLT \cite{nayef2017icdar2017}, Total-Text \cite{ch2017total}, CTW1500 \cite{yuliang2017detecting}, MSRA-TD500 \cite{yao2012detecting}, and achieve state-of-the-art results. 
While fine-tuning is a common practice in recent works \cite{zhou2017east,sheng2021centripetaltext}, the proposed model is directly trained using a combination of datasets without fine-tuning on each individual target dataset. The unified detector is the first end-to-end model that achieves state-of-the-art performance on the text detection task and simultaneously recovers important text layout information.

In conclusion, our core contributions are as follows:

\begin{compactitem}
    \item We propose the task of unified text detection and layout analysis, bringing together two tasks that have been studied independently, yet are intrinsically connected. 
    \item A new high quality dataset with hierarchical text annotations is introduced to facilitate research on this task.
    \item We propose an end-to-end unified model, which outperforms competitive multi-stage baselines that treat the two tasks separately.
    \item Our model, which is free of complex post-processing, achieves state-of-the-art results on multiple challenging public text detection benchmarks.
\end{compactitem}

% ----- ----- ----- ----- ----- Related Works ----- ----- ----- ----- ----- 
\section{Related Works}
\label{sec:rel}

\subsection{Scene text and documents datasets}
There have been a variety of scene text datasets and document datasets. Scene text datasets range from straight text \cite{karatzas2015icdar} to curved text\cite{ch2017total,yuliang2017detecting}, sparse text to dense text\cite{tang2019seglink,Singh_2021_CVPR}, monolingual text  \cite{karatzas2015icdar,ch2017total} to multilingual text\cite{nayef2017icdar2017,nayef2019icdar2019}, word level to line level\cite{yuliang2017detecting,yao2012detecting}, narrower image domain to broader image domain\cite{Singh_2021_CVPR,krylov2021open}, varying in characteristics. 
However, these datasets only focus on the retrieval of individual words or text lines.
There are also datasets that provide additional higher-level annotations for text based VQA\cite{Singh_2021_CVPR} and image captioning\cite{xu2021towards}. However, they focus on specific tasks and do not analyze the layout of text, which has universal usage in downstream tasks.
Document datasets\cite{clausner2015enp,clausner2017icdar2017,antonacopoulos2009realistic,zhong2019publaynet,li2020end} only provide annotations for layout analysis without labeling the atomic entities i.e. words. 
Furthermore, these datasets only contain scanned or digital documents for a specific domain such as academic papers\cite{zhong2019publaynet} and historical newspapers\cite{clausner2015enp}. 
The text reading order dataset\cite{li2020end} only contains images that have well-defined reading order, such as product labels and instruction manuals, and thus is not general.
The proposed dataset is the first dataset that allows joint detection and layout analysis for general natural images.

\subsection{Scene text detection}
Recently, scene text detection research\cite{long2021scene} has mainly focused on the representation of irregularly shaped text and post-processing method that recovers the text contours from geometric attributes such as word or character center regions, pixel level orientation, and radius of the text\cite{long2018textsnake,baek2019character,wang2019shape,liao2020mask,Dai_2021_CVPR,sheng2021centripetaltext}. 
The custom representation for text complicates the label generation process and post-processing, such as semi-supervised and iterative generation of character center regions\cite{baek2019character}, boundary shrinking and recovery\cite{liao2020mask} with Vatti clipping\cite{vatti1992generic}, and polygonal non-maximum suppression\cite{Dai_2021_CVPR}.
Raisi et al. \cite{raisi2021transformer} introduce the end-to-end detector DETR\cite{carion2020end} to detect text using rotated bounding boxes, but it does not handle curved text.
Besides, these works only provide solutions to the task of text detection.
Conversely, our research works on the unification of text detection and layout analysis with an end-to-end neural network that greatly simplifies the whole pipeline.

\subsection{Layout analysis}
Driven by the success of object detection\cite{ren2015faster,he2017mask} and semantic segmentation \cite{long2015fully,chen2017deeplab} in images, layout analysis in documents is also framed as detection and segmentation tasks in some works\cite{schreiber2017deepdesrt,lee2019page,zhong2019publaynet}, where detector models are trained to detect semantically coherent text blocks as objects. 
These methods fail to produce word or line level detections and can only be used in company with standalone text detectors, increasing the complexity of the pipeline. 
Another branch of work\cite{wang2022post} takes a hierarchical view and apply graph-based models on the finest granularity, i.e. individual words, to analyze the layout. 
All of these prior arts have mainly focused on document datasets. Unlike these works, we introduce layout analysis into scene text domain and propose an end-to-end unified model.

% ----- ----- ----- ----- ----- Dataset intro ----- ----- ----- ----- ----- 

\begin{table}
  \centering
  \scalebox{0.65}{
\begin{tabular}{cp{1.0cm}p{1.0cm}p{1.0cm}cp{0.5cm}p{0.5cm}p{1.2cm}}
\hline
\multirow{2}{*}{Dataset} & \multicolumn{3}{c}{\#Img}                            & \multirow{2}{*}{\begin{tabular}[c]{@{}c@{}}\#Word\\ (avg/total)\end{tabular}} & \multicolumn{3}{c}{Ann Level}  \\ \cline{2-4} \cline{6-8} 
                         & Train            & Val             & Test            &                                                                               & Word & Line & Paragraph \\ \hline
IC15  \cite{karatzas2015icdar}& 1,000            & 0               & 500             & 4.4/6.5K                                                                    & \checkmark    &            &            \\
Total-Text\cite{ch2017total}               & 1,255            & 0               & 300             & 7.4/11K                                                                     & \checkmark    &            &            \\
CTW1500\cite{yuliang2017detecting}                  & 1,000            & 0               & 500             & 6.7/10K                                                                     &      & \checkmark          &            \\
MSRA-TD500 \cite{yao2012detecting}              & 300              & 0               & 200             & 6.9/3.5K                                                                    &      & \checkmark          &            \\ \hline
IC17 MLT$^\star$\cite{nayef2017icdar2017}                 & 7,200            & 1,800           & 9,000           & 9.5/85K                                                                    & \checkmark    &            &            \\
IC19 MLT$^\star$ \cite{nayef2019icdar2019} & 10,000           & 0               & 10,000          & 8.9/89K                                                                    & \checkmark    &            &            \\
IC19 LSVT$^\star$ \cite{sun2019icdar}                & 30,000           & 0               & \textbf{20,000} & 8.1/243K                                                                   &      & \checkmark          &            \\
IC19 ArT$^\star$ \cite{chng2019icdar2019}               & 5,603            & 0               & 4,563           & 8.9/50K                                                                    & \checkmark    &            &            \\ \hline
TextOCR \cite{Singh_2021_CVPR}                 & 21,778           & 3,124           & 3,232           & 32.1/903K                                                                   & \checkmark    &            &            \\
Intel OCR  \cite{krylov2021open}              & \textbf{191,059} & \textbf{16,731} & 0               & 10.0/\textbf{2.1M}                                                          & \checkmark    &            &            \\ \hline
\begin{tabular}[c]{@{}c@{}}\textbf{HierText}\end{tabular}           & 8,281            & 1,724           & 1,634           & \textbf{103.8}/1.2M                                                         & \checkmark    & \textbf{\checkmark} & \textbf{\checkmark} \\ \hline
\end{tabular}}
  \caption{HierText v.s. other datasets. Our dataset is characterized by high text density and hierarchical annotations. Datasets marked with $^\star$ do not provide test set annotations. The train and validation sets were used to count the number of words.}
  \label{tab:stats}
  \vspace{-4mm}
\end{table}

\section{Hierarchical Text Dataset (HierText)}
\label{sec:dataset}

\subsection{Data collection}

Images in HierText are collected from the Open Images v6 dataset \cite{kuznetsova2020open}. We scan Open Images using a public commercial OCR engine, \textit{Google Cloud Platform Text Detection API (GCP)}\footnote{\url{https://cloud.google.com/vision/docs/ocr}}, to search for images with text. 
We filter out images: a) with few detected words, b) low recognition confidence and c) with non-English dominant text. Finally, we randomly sample a subset from the remaining images to construct our dataset. $11639$ images are obtained and further splitted into \textit{train}, \textit{validation}, and \textit{test} set. HierText images are of higher resolution with their long side constrained to $1600$ pixels compared to previous datasets based on Open Images \cite{Singh_2021_CVPR,krylov2021open} that are constrained to $1024$ pixels, resulting in more legible text.

We annotate these images in a hierarchical way \cite{haralick1994document}. We first annotate word locations with polygons. Legible words are also transcribed regardless of their language. The top-left corner and the orientation of the polygon define the word's reading direction. Then words are grouped to text lines. Paragraphs are firstly annotated using polygon and then text lines and words are associated with the corresponding polygon based on their binary mask intersection. As a result, we obtain a tree structure annotation hierarchy. Note that, the clustering of words into lines and lines into paragraphs are relatively low-cost, since precise pixel level annotation is not required.

\noindent \textbf{Coverage check}: We check the cross-dataset coverage between HierText and the other two text datasets from Open Images, i.e. TextOCR \cite{Singh_2021_CVPR} and Intel OCR \cite{krylov2021open}. Only $1.5\%$ of our images are in TextOCR, and $3.6\%$ in Intel OCR. We also ensure that our training images are not in the validation or test set images of TextOCR and Intel OCR, or vice versa.

\begin{figure}[t]
 \centering
  \begin{subfigure}[t]{.495\linewidth}
      \centering
      \includegraphics[width=1\linewidth]{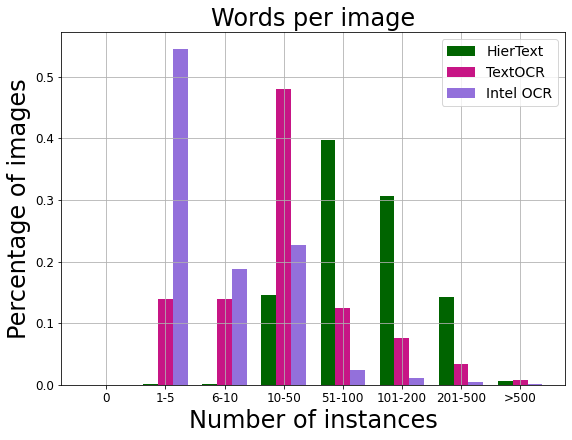}
      \caption{Text density distribution.}
      \label{fig:distribution}
  \end{subfigure}
  %\hfill
  \begin{subfigure}[t]{.495\linewidth}
      \centering
      \includegraphics[width=\linewidth]{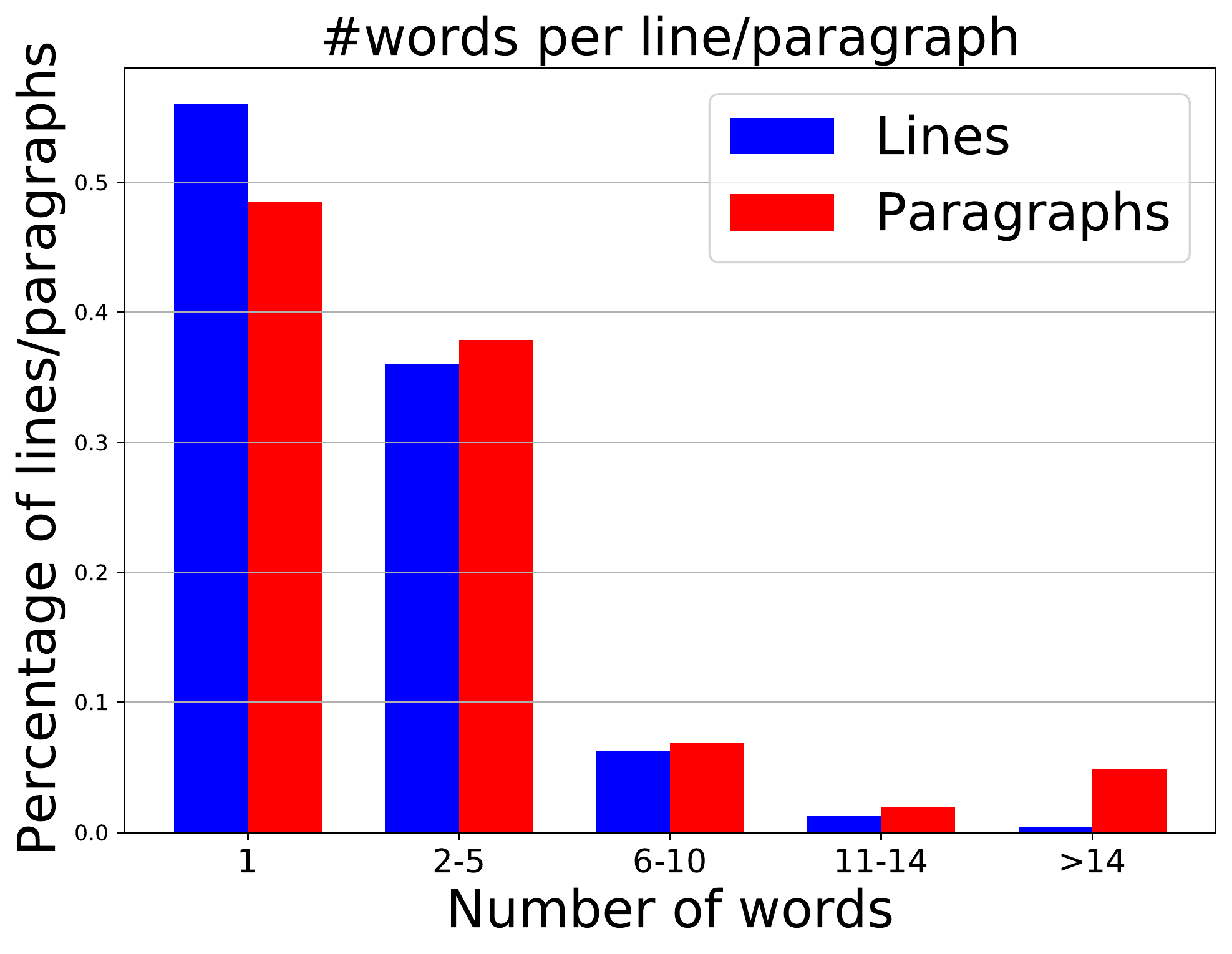}
      \caption{Distribution of numbers of words in each line and paragraph in HierText.}
      \label{fig:words_per_line_paragraph}
  \end{subfigure}\\
  \begin{subfigure}[b]{\linewidth}
      \includegraphics[width=\linewidth]{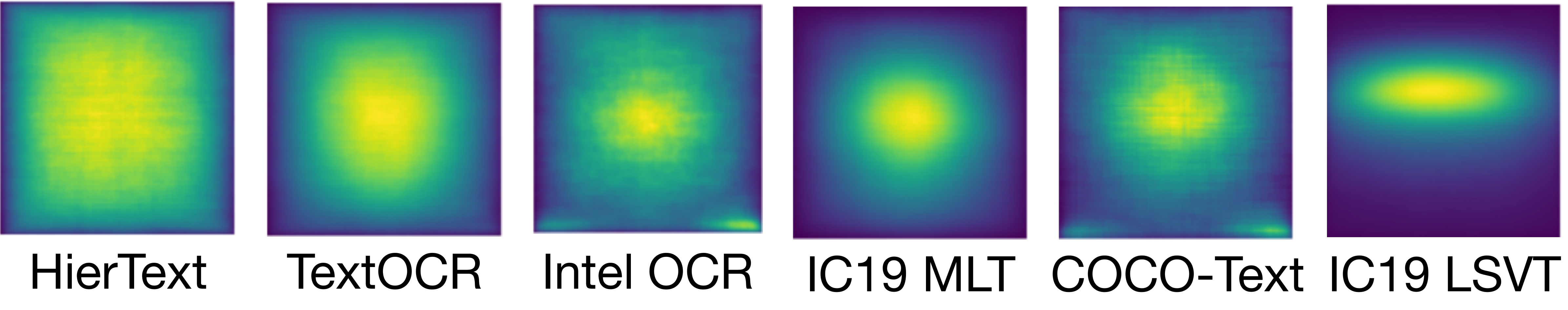}
      \caption{Text spatial distributions.}
      \label{fig:location_distributions}
  \end{subfigure}
  \caption{\textbf{Statistics of HierText dataset.} Compared to other datasets, our dataset has higher word density per image and more uniformly distributed text.}
  \vspace{-3mm}
\end{figure}

\subsection{Dataset characteristics}
Table~\ref{tab:stats} compares statistics between HierText and other datasets. HierText has $103.8$ words per image on average; approximately $3$ times the text density of  the second densest dataset, i.e. TextOCR\cite{Singh_2021_CVPR}. 
Even though HierText has fewer images than TextOCR, it contains more legible words. Finally, HierText is the only dataset that provides hierarchical annotations. 
Fig. \ref{fig:distribution} shows that HierText represents a different domain of images from existing public datasets. It has a large proportion of high text density images. While Intel OCR\cite{krylov2021open} has the largest number of images and some coverage of images with more than $100$ words, HierText contains more of them in terms of absolute number: $5.3K$ v.s. $3.4K$. Fig.\ref{fig:location_distributions} illustrates that the spatial distribution of text is also more uniform in HierText. In other datasets, text tend to be located in the center of the images. The distribution of the number of words in each line and paragraph is shown in Fig. \ref{fig:words_per_line_paragraph}. A significant proportion of lines and paragraphs have more than one word making the layout analysis a challenging problem. 

Overall, we demonstrate that the proposed HierText dataset has unique characteristics and captures an uncovered domain from other datasets. Additionally, it enables research into unified text detection and layout analysis. 

\subsection{Task and evaluation protocol}
\noindent \textbf{Tasks}: There are two task categories for HierText dataset. The first category involves the instance segmentation of text at word or line levels. 
Conceptually, word level and line level outputs are interchangeable since modern text recognition systems \cite{shi2018aster,lu2021master,Cheng_2017_ICCV} are highly effective with both type of input image patches. 
For the second task of layout analysis, we also frame it as an instance segmentation task by treating each text cluster, i.e. ``paragraph'', as one object instance, following previous works\cite{zhong2019publaynet}.
The ground-truths for text lines and paragraphs are defined as the union of pixel-level masks of the underlying word level polygons. 

A candidate method for the unified detection and layout analysis task should produce text entity detection results at either word or line level, and also the grouping of these entities into paragraphs.

\noindent \textbf{Evaluation}: To evaluate these tasks as instance segmentation, we use the recently proposed Panoptic Quality (PQ) metric \cite{kirillov2019panoptic} as the main evaluation metric:

\begin{equation}
  PQ = \frac{\sum_{(p, g)\in TP}IoU(p, g)}{|TP| + \frac{1}{2}|FP| + \frac{1}{2}|FN|}
  \label{eq:pq-def}
\end{equation}

\noindent where $TP,~FP,~FN$ represent  true positives, false positives, and false negatives respectively. 
Mathematically, PQ equals to the product of \textit{ICDAR15 \cite{karatzas2015icdar} style F1 score \cite{everingham2015pascal}} and \textit{average IoU of all TP pairs}. The motivation for a segmentation metric is that text entities are sensitive to missing or superfluous pixels which result in missing or unexpected characters in recognition. 
Although there have been recent works \cite{shi2017icdar2017,liu2019tightness,lee2019tedeval} investigating the scoring of text detection, they do not generalize to complex geometric entities like text lines and paragraphs. PQ metric treats word, line and paragraph segmentation tasks in a uniform way. Therefore, we adopt PQ metric for the evaluation of all tasks for its simplicity and ubiquity.

% ----- ----- ----- ----- ----- Method ----- ----- ----- ----- ----- 
\section{Methodology}
\label{sec:method}
\subsection{Unified detector}
We propose an end-to-end model to perform unified scene text detection and layout analysis. We term it \textit{Unified Detector}. 
It is designed to produce (1) a set of text detection masks and (2) the clustering of these detections simultaneously without complex post-processing. 

\noindent \textbf{End-to-end text detection}: Inspired by recent advances in end-to-end object detection and panoptic segmentation\cite{carion2020end,wang2021max}, we represent the detection of text as producing a fixed number of $N$ softly exclusive masks $\{\hat{m}_i\}_{i=1}^N$ and $N$ binary classifications $\{\hat{y}_i\}_{i=1}^N$. 
The masks satisfy $\sum_{i=1}^N \hat{m_i} = \mathbbm{1}^{H\times W}$. 
The binary classification $\hat{y}_i$ denotes the probability of the $i-$th mask being a text object. 
This representation is suitable for text of arbitrarily shape and can accurately capture both word and line level detections.

\noindent \textbf{Unified layout analysis}: Unified detector analyzes the layout and performs text clustering by producing an affinity matrix: $\hat{A}\in [0, 1]^{N\times N}$. 
Entry $\hat{A}_{i,j}$ in this matrix represents the probability of text represented by $\hat{m_i}$ and $\hat{m_j}$ belonging to the same semantic/paragraph group.

\noindent \textbf{Inference}: The inference of unified detector is straightforward. We first obtain text detection results by applying argmax on the masks to assign each pixel to one text object. Then, we remove low-confidence pixels. As a result, for the $i-$th object, the final mask is represented as: 

\begin{equation}
  m'_{i,x,y} =  \mathbbm{1}(i=\text{argmax}_j[\hat{m}_{j,x,y}]\ \text{and}\  \hat{m}_{i,x,y} > t_m)
  \label{eq:mask-inf}
\end{equation}

\noindent where $t_m$ is the threshold for pixel's confidence. We further filter text instances by applying a threshold $t_c$ on the binary classification score $\hat{y}_i$. For layout analysis inference, we cluster a pair of text instances if their affinity score $\hat{A}_{i,j}$ is above a threshold, denoted as $t_A$. A union-find algorithm is utilized to merge these connected nodes into clusters.

\subsection{Model architecture}
\begin{figure}[t]
  \centering
   \includegraphics[width=1.0\linewidth]{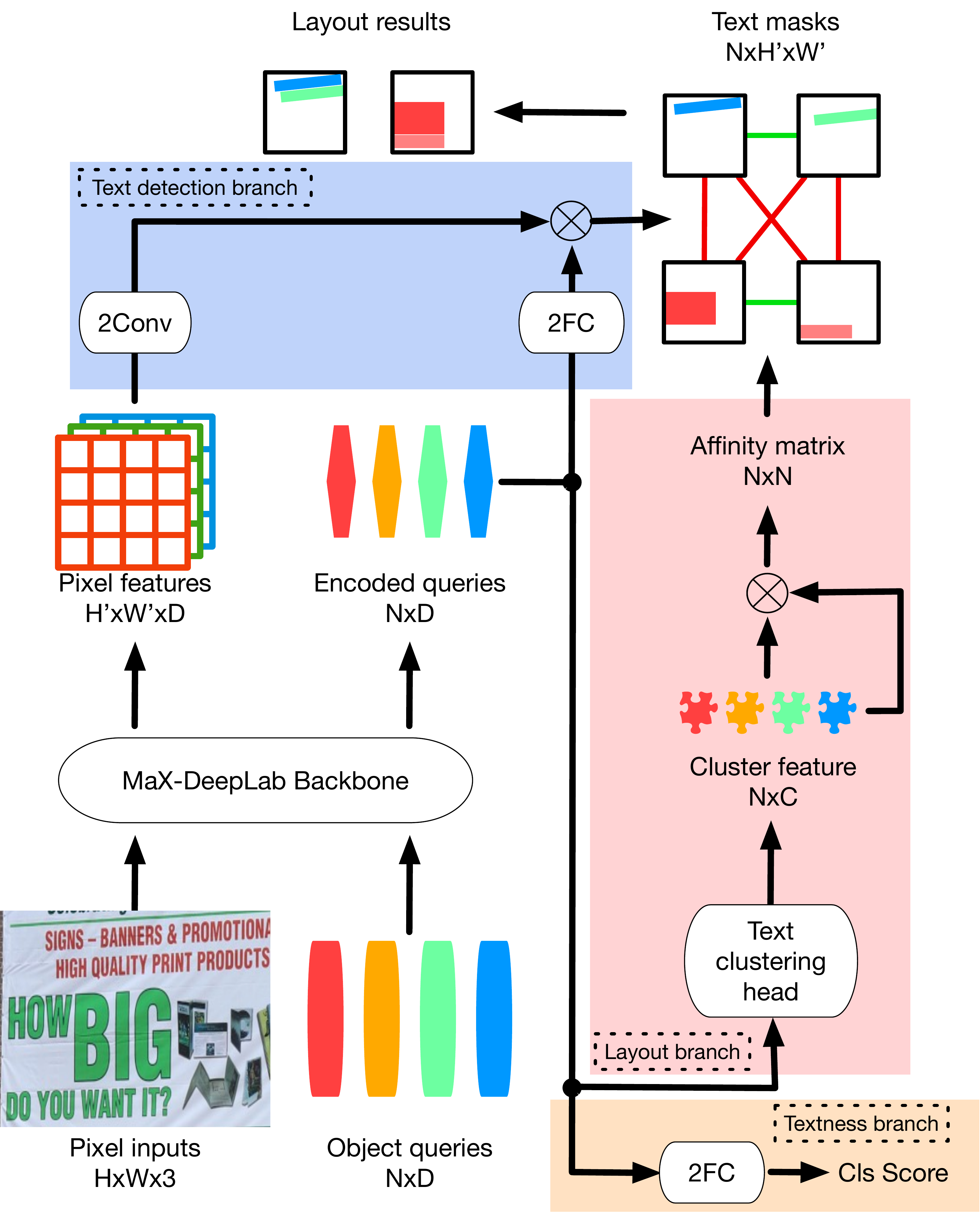}

   \caption{Illustration of our approach. Our method produces mask outputs for the text detection task and an affinity matrix for layout analysis task in a unified way. The text detection branch produces an $N\times H\sp{\prime}\times W\sp{\prime}$ tensor that represents the $N$ softly exclusive masks. The layout analysis branch produces an $N\times N$ affinity matrix that models the pairwise relationship of the predicted masks. The {\color{green}green} links in the top right suggest clustering of the text instances, while the {\color{red}red} links indicate the opposite. The binary classification score produced by textness branch is used to filter out non-text objects from object queries.}
   \label{fig:model}
\end{figure}

The architecture of the proposed unified detector is illustrated in Fig. \ref{fig:model}. 
Our unified detector is based on the recent Max-DeepLab \cite{wang2021max} end-to-end panoptic segmentation framework. 
In this framework, we augment the input pixels with a set of $N$ learned object queries that are $D-$dimensional. 
Then we feed the pixels and object queries into a transformer-based encoder, the MaX-DeepLab backbone, in which the bidirectional communication between pixels and object queries allows the model to encode text instances in each of the object queries. With the encoded queries and pixel features, the \textit{text detection branch} produces the text mask output, $\{\hat{m}_i\}_{i=1}^N$. The \textit{layout branch} produces the affinity matrix $\hat{A}\in [0, 1]^{N\times N}$ for the relations between each pair of text instances. A third branch produces the binary classification scores $\{\hat{y}_i\}_{i=1}^N$.

\noindent \textbf{Backbone}:
The MaX-DeepLab \cite{wang2021max} backbone is composed of an alternating stack of hourglass \cite{newell2016stacked} style CNNs and the proposed dual-path transformer. The Hourglass style \cite{newell2016stacked} CNNs are applied to pixel features. They encode features from coarse to fine resolutions iteratively and thus can produce high resolution features.
The dual-path transformer \cite{wang2021max} allows bidirectional communication between pixel features and the learnable object queries. It enables attention within pixel space and interaction among object queries. 
This makes it possible to encode long-range information in pixel features, and allows object queries to locate and retrieve text objects exclusively from pixels. 
The MaX-DeepLab produces output at $\frac{1}{4}$ resolution of the input, i.e. $(H\sp{\prime}, W\sp{\prime}) = (\frac{H}{4}, \frac{W}{4})$.
We urge readers to refer to the original paper \cite{wang2021max} for full details.

\noindent \textbf{Text detection branch}:
The text detection branch takes the outputs of the MaX-DeepLab backbone and produces the text mask outputs. 
Two fully-connected layers produce mask queries from the encoded queries, denoted as $f\in \mathbb{R}^{N\times D}$. 
Similarly, two convolutional layers produce normalized pixel features, denoted as $g\in \mathbb{R}^{D\times H'\times W'}$.
The text mask prediction is the inner product of $f$ and $g$:

\begin{equation}
  \hat{m} = \text{softmax}_{N} (f \cdot g) \in \mathbb{R}^{N\times H'\times W'}
  \label{eq:mask-head}
\end{equation}

\noindent \textbf{Layout branch}: Layout branch takes the encoded queries from the backbone as the sole input. 
In order to separate layout features from text detection features, we apply an extra projection head for cluster embedding projection. 
For this projection head, we adopt a 3-layered multi-head self-attention layer \cite{vaswani2017attention} to obtain the normalized layout features, denoted as $h\in \mathbb{R}^{N\times C}$. We apply inner product of the layout features followed by a sigmoid function with temperature $\tau$ to get the affinity matrix:

\begin{equation}
  \hat{A}[i, j] = \frac{1}{1+e^{-(\frac{h_{i, :} h_{j, :}^T}{\tau})}}
  \label{eq:layout-head}
\end{equation}

\noindent \textbf{Textness branch}: The textness branch applies another $2$-layered fully connected layers and a sigmoid function to produce the binary classification scores $\{\hat{y}_i\}_{i=1}^N$.

\subsection{Training targets} 
\label{sec:loss}
Unified detector enables end-to-end training for both the scene text detection task and the layout analysis task. 
The key ingredient is to perform a bipartite matching between prediction and groundtruth since our model produces an unordered set of outputs.
We first describe the matching between prediction and groundtruth of the detection task and the metric we use. Then we show the joint optimization of our unified detector for both tasks.

\noindent\textbf{Text matching}:
We adopt the PQ-style similarity score proposed in MaX-DeepLab \cite{wang2021max}. For a pair of prediction ($\hat{m}_i, \hat{y}_i$) and groundtruth ($m_j, y_j$), the score is defined as:

\begin{equation}
  \text{sim}(i, j) = [\hat{y}_iy_j + (1-\hat{y}_i)(1-y_j)]\times \text{Dice}(\hat{m}_i, m_j)
  \label{eq:loss-sim}
\end{equation}

\noindent where $\text{Dice}(\hat{m}_i, m_j)$ denotes the Dice coefficient \cite{milletari2016v} between the pair of masks. It measures mask similarity. This score considers both the classification score and mask score. 

The goal of matching is to find a permutation of $N$ elements $\sigma\in\mathfrak{G}_N$ to maximize the total similarity between predictions and groundtruths:

\begin{equation}
  \hat{\sigma} = \text{argmax}_{\sigma} \sum_{i=1}^{N}\text{sim}(i, \sigma(i))
  \label{eq:loss-match}
\end{equation}

\noindent Following previous works \cite{wang2021max,carion2020end}, we solve this optimal assignment problem  with the Hungarian algorithm \cite{kuhn1955hungarian} on the fly during training.

\noindent\textbf{Text detection loss}: The training target for text detection is adopted from MaX-DeepLab \cite{wang2021max}:

\begin{multline}
  \mathcal{L}_{det} = \frac{1}{N}\sum_{i=1}^N \{(1 - \alpha)(1 - y_{\sigma(i)})[-\text{log}(1-\hat{y}_i)] \\
  + \alpha y_{\sigma(i)}[
  -\Ddot{\hat{y_i}}\text{Dice}(\hat{m}_i, m_{\sigma(i)}) 
  -\Ddot{\text{Dice}}(\hat{m}_i, m_{\sigma(i)})\text{log}(\hat{y}_i)
  ]
  \}
  \label{eq:loss-det}
\end{multline}

\noindent where dotted variables $\Ddot{\hat{y_i}}$ and $\Ddot{\text{Dice}}(\hat{m}_i, m_{\sigma(i)})$ denote constant weights and gradients do not pass through them. $\alpha$ is a balancing factor between positive and negative samples.

\noindent\textbf{Layout analysis loss}: We first define the ground-truths for output of layout analysis branch. Each text instance comes with a text cluster ID, denoted as $\{c_i\}_{i=1}^N$. This is part of the annotations of the proposed HierText dataset. The groundtruth affinity matrix can be intuitively defined as:

\begin{equation}
  A[i,j] = \mathbbm{1}(c_i == c_j)
  \label{eq:loss-aff-def}
\end{equation}

\noindent Then, the layout analysis loss can be computed as:

\begin{multline}
  \mathcal{L}_{lay} = \sum_{i=1}^N\sum_{j=1}^Ny_{\sigma(i)}y_{\sigma(j)}\{
  \alpha_L w_p A_{\sigma(i), \sigma(j)}[-\text{log}(\hat{A}_{i, j})] \\
  + (1 - \alpha_L) w_n (1 - A_{\sigma(i), \sigma(j)})[-\text{log}(1 - \hat{A}_{i, j})]\}
  \label{eq:loss-aff}
\end{multline}

\noindent where $w_p = [\sum_{i=1}^N\sum_{j=1}^Ny_{\sigma(i)}y_{\sigma(j)}
  A_{\sigma(i), \sigma(j)}]^{-1}$, $w_n = [\sum_{i=1}^N\sum_{j=1}^Ny_{\sigma(i)}y_{\sigma(j)}
  (1-A_{\sigma(i), \sigma(j)})]^{-1} $, and $\alpha_L$ is a balancing factor.

The final training target is the weighted sum of the text detection loss $\mathcal{L}_{det}$, the layout analysis loss $\mathcal{L}_{lay}$. We also find it useful to incorporate the semantic segmentation loss $\mathcal{L}_{seg}$ and instance discrimination loss $\mathcal{L}_{ins}$ as defined in MaX-DeepLab \cite{wang2021max}. As a result, the model is jointly optimized for the following loss function:

\begin{equation}
  \mathcal{L} = \lambda_1\mathcal{L}_{det} + \lambda_2\mathcal{L}_{lay} + \lambda_3\mathcal{L}_{seg} + \lambda_4\mathcal{L}_{ins}
  \label{eq:loss}
\end{equation}

\noindent where $\lambda_{1,\ldots,4}$ are weighting factors.

% ----- ----- ----- ----- ----- Exp1 ----- ----- ----- ----- ----- 
\section{Experiments}
\label{sec:exp}

In this section, we set up experiments to evaluate our proposed unified detector in a comprehensive way. First, we compare our method with competitive baselines. We show that the unified detector achieves better performance. We also perform thorough ablation studies to analyze the design selections of the proposed approach. Finally, we train and evaluate the unified detector on public datasets for the sole task of scene text detection, verifying the effectiveness of the text detection branch.

\subsection{Baselines}

The task of unified detection and layout analysis largely remains untouched in the academic literature, despite the incredible progress of scene text detection methods and increasing number of layout analysis algorithms. 
We therefore carefully select the following baselines representing non-end-to-end methods:

\noindent \textbf{Commercial solution}: The \textit{GCP} API, as mentioned above, is a  commercial solution that produces text detection and recognition results at word, line and paragraph level. 

\noindent \textbf{GCN Post-Processing}: The GCN\cite{kipf2016semi} based post-processing method (GCN-PP) \cite{wang2022post} applies the GCN on text line bounding boxes to cluster lines into paragraphs.

\begin{figure}[t]
  \centering
   \includegraphics[width=1.0\linewidth]{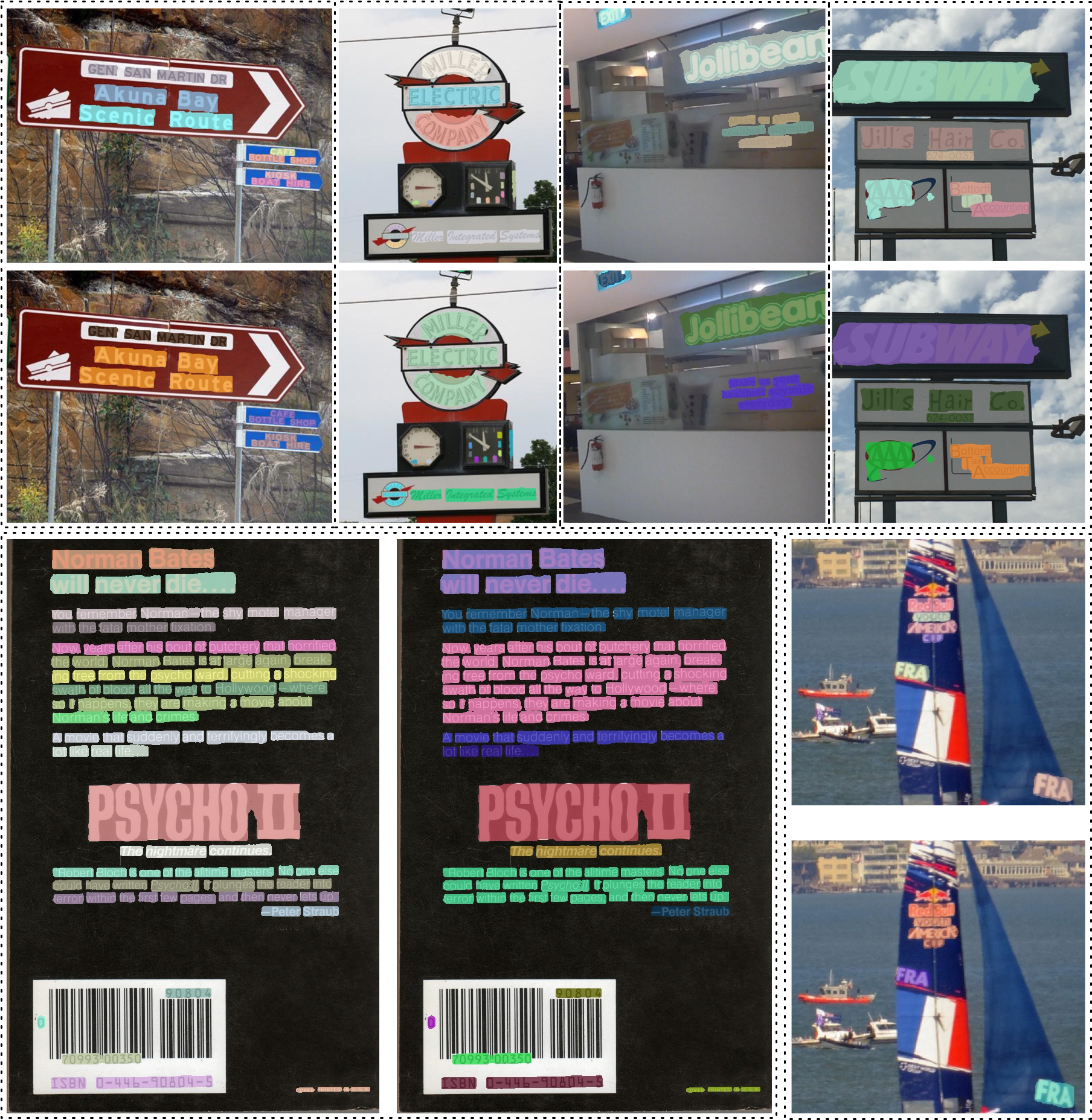}

   \caption{Outputs of the unified detector trained on HierText. Images are sampled from the val and test set of HierText, Total-Text, CTW1500, IC15, ICDAR17 MLT, and MSRA-TD500. In each pair of images bounded by dotted boxes, the image on the \textbf{top or left} visualizes \textbf{text detection results}. The image on the \textbf{bottom or right} visualizes results of \textbf{layout analysis}. These results demonstrate that our unified detector is able to detect arbitrarily shaped text and produce text clusters regardless of the variability in shapes, fonts, colors and backgrounds.}
   \label{fig:demo}
\end{figure}

\noindent \textbf{Object detection baselines}:  PubLayNet\cite{zhong2019publaynet} formulates the layout analysis as an instance segmentation task predicting text clusters as pixel masks. Following this work, we build a baseline using Mask R-CNN\cite{he2017mask} as in \cite{zhong2019publaynet} that produces text cluster masks. Each such mask represents one text cluster. The layout analysis is performed by assigning each detected text entity (word or line) to the text cluster with the maximum area of intersection. Since this model does not produce word or line level detections, it is used in combination with a text entity detection model as specified in Sec. \ref{exp:unified}. This two-stage baseline is dubbed \textit{Mask-RCNN-Cluster}. 
Similarly, we build \textit{MaX-DeepLab-Cluster} using MaX-DeepLab\cite{wang2021max}, which represents a more competitive method with state-of-the-art advance in instance object segmentation task.

\begin{table}[]
\scalebox{0.65}{
\begin{tabular}{ccccc}
\hline
                         &                                                                                                         &                                                                                    & \begin{tabular}[c]{@{}c@{}}Text line\\ Detection\end{tabular} & \begin{tabular}[c]{@{}c@{}}Layout\\ Analysis\end{tabular} \\ \cline{4-5} 
\multirow{-2}{*}{Method} & \multirow{-2}{*}{\begin{tabular}[c]{@{}c@{}}Text detection\\ branch\end{tabular}}                       & \multirow{-2}{*}{\begin{tabular}[c]{@{}c@{}}Layout analysis\\ branch\end{tabular}} & PQ                                                            & PQ                                                        \\ \hline
GCP API                  & unknown                                                                                                 & unknown                                                                            & 56.17                                                         & 46.33                                                     \\ \hline
GCN-PP                   &                                                                                                         & GCN                                                                                &                                                               & 50.10                                                     \\ \cline{1-1} \cline{3-3} \cline{5-5} 
Mask-RCNN-Cluster        &                                                                                                         & Mask R-CNN \cite{he2017mask}                                                                         &                                                               & {\color[HTML]{000000} 51.67}                              \\ \cline{1-1} \cline{3-3} \cline{5-5} 
MaX-DeepLab-Cluster      &                                                                                                         & MaX-DeepLab  \cite{wang2021max}                                                                      &                                                               & {\color[HTML]{000000} 52.52}                              \\ \cline{1-1} \cline{3-3} \cline{5-5} 
\textbf{Unified Detector}         & \multirow{-4}{*}{\begin{tabular}[c]{@{}c@{}}Text detection\\ branch of\\ unified detector\end{tabular}} & \begin{tabular}[c]{@{}c@{}}Layout branch of \\ unified detector\end{tabular}       & \multirow{-4}{*}{\textbf{62.23}}                              & {\color[HTML]{000000} \textbf{53.60}}                     \\ \hline
\end{tabular}
}
\caption{Results of text detection and layout analysis on HierText test set. The last row represents our end-to-end unified scene text detection and layout analysis model.}
\label{tab:unified}
% \vspace{-4mm}
\end{table}

\subsection{Experimental settings}
\label{exp:unified}
% training setting, hyper-parameters etc.
\noindent \textbf{Unified Detector}: We use the DeepLab2 \cite{deeplab2_2021} library for the implementation of the MaX-DeepLab part of our method. We use the MaX-DeepLab-S backbone, with an input size of $1024\times1024$. The number of object query is set to 384, due to the high density of text in our dataset. Query dimensions are $D=256$ and $C=128$ respectively. In our main experiments, we only use HierText as training data. The models are trained on 128 TPUv3 cores with a batch size of 256 for 100K steps, AdamW\cite{loshchilov2017decoupled} optimizer with weight decay rate of 0.05, and cosine learning rate starting from $10^{-3}$. The weights for PQ-loss, layout analysis loss, instance discrimination loss, and semantic segmentation loss are $3.0$, $1.0$, $1.0$, $1.0$ respectively. 
The balancing factors are set as $\alpha=0.5$ and $\alpha_L=0.5$.
During inference, we filter out text masks that have less than $32$ pixels or less than $t_c=0.5$ in confidence. We also use $t_m=0.4$ to filter out low confidence pixels. For text clustering, we use a threshold of $t_A=0.5$ on the affinity matrix. In our main experiments, the text detection branch of unified detector is trained to detect text lines as opposed to words. Note that most of these hyper-parameters follow the original settings of MaX-DeepLab.

\noindent \textbf{Baselines}: For Mask-R-CNN-Cluster, we use the implementation from the public TF-Vision repository\footnote{\url{https://github.com/tensorflow/models/tree/master/official/vision/beta}}. Input size is set to $1024\times1024$. For MaX-DeepLab-Cluster, we follow the same hyper-parameter and training settings of our unified detector for fair comparison. For GCN-PP, we follow the settings in \cite{wang2022post} to train the line clustering model.
As mentioned above, these methods can only perform layout analysis based on detected text entities. Therefore, we pair these three baselines with the text detection branch of our unified detector for fair comparison.

\subsection{Main Results}

We evaluate our method and compare with baselines as detailed above. Results are summarized in Tab. \ref{tab:unified}.
Compared with other standalone text clustering methods including GCN-based and detection-based ones, our end-to-end unified approach achieves better layout analysis performance by a considerable margin of $1.08\%$ in PQ score. 
Note that, these baseline methods are applied on the outputs of the text detection branch of unified detector. Therefore, the only difference is in the layout analysis method. 
This shows that the built-in end-to-end text clustering module of unified detector is more effective and better than standalone baseline modules. Note that the baselines are two stage approaches that require almost double the computational resources.
For text detection, our system achieves higher performance than the GCP API ($62.23$ v.s. $56.17$).

We also demonstrate results on images from various domains, as shown in Fig. \ref{fig:demo}. 
The proposed method is able to work on various layouts, including text clusters with curved text and with non-uniform fonts and colors.

\subsection{Ablation studies}

\begin{table}[]
\centering
\scalebox{0.8}{\begin{tabular}{ccc}
\hline
\#Obj query            & Method                  & \begin{tabular}[c]{@{}c@{}}Layout analysis\\(PQ)\end{tabular} \\ \hline
\multirow{2}{*}{128} & Unified-Detector-Word & 34.38        \\
                     & Unified-Detector-Line & 51.48        \\ \hline
\multirow{2}{*}{256} & Unified-Detector-Word & 36.71       \\
                     & Unified-Detector-Line & 52.50        \\ \hline
\multirow{2}{*}{384} & Unified-Detector-Word & 39.11        \\
                     & Unified-Detector-Line & \textbf{53.60}        \\ \hline
\end{tabular}}
\caption{Comparison between word-based and line-based unified detector. Results demonstrate that line-based unified detector outperforms word-based unified detector consistently with different numbers of object queries.}
\label{tab:word-line}
\end{table}

\begin{table}[]
\scalebox{0.75}{\begin{tabular}{c|cclclc}
\hline
& \multicolumn{5}{c}{Text Line Detection}&\\
\multirow{-2}{*}{\begin{tabular}[c]{@{}c@{}}Balancing\\method\end{tabular}} & P & R & \multicolumn{1}{c}{F} & T & \multicolumn{1}{c}{PQ} & \multirow{-2}{*}{\begin{tabular}[c]{@{}c@{}}Layout analysis\\ (PQ)\end{tabular}} \\ \hline
Vanilla & \multicolumn{1}{l}{75.34} & \multicolumn{1}{l}{75.02} & 75.18 & \multicolumn{1}{l}{77.27} & 58.10 & 50.04\\ \hline
\begin{tabular}[c]{@{}c@{}}$\alpha$-balanced loss\\  ($\alpha=0.25$) \end{tabular}  & 76.32 & 75.20 & \multicolumn{1}{c}{75.76} & 77.57 & \multicolumn{1}{c}{\textbf{58.76}}  & {\color[HTML]{000000} \textbf{51.48}} \\  \hline
focal loss & 75.16 & 74.58 & 74.87 & 77.38 & 57.94 & {\color[HTML]{000000} 45.22} \\ \hline
\end{tabular}}
\caption{The effect of balancing factor in text clustering loss on text and layout metrics.}
\label{tab:ablation-loss}
% \vspace{-4mm}
\end{table}

% ----- ----- ----- ----- ----- Exp2 ----- ----- ----- ----- ----- 

\begin{table*}[]
\centering
\scalebox{0.8}{\begin{tabular}{@{}cccccccccccccccc@{}}

\toprule
\multirow{3}{*}{Method} & \multirow{3}{*}{Venue} & \multicolumn{2}{c}{Training Data} & \multicolumn{6}{c}{Word Detection} & \multicolumn{6}{c}{Line Detection} \\ \cmidrule(l){3-16} 
 & & \multirow{2}{*}{Pub} & \multirow{2}{*}{HierText} & \multicolumn{3}{c}{ICDAR 17 MLT} & \multicolumn{3}{c}{Total-Text} & \multicolumn{3}{c}{CTW1500} & \multicolumn{3}{c}{MSRA-TD500} \\ \cmidrule(l){5-16} 
 & & & & P & R & F & P & R & F & P & R & F & P & R & F \\ \midrule
 
CRAFT \cite{baek2019character} & CVPR19 & \checkmark & & 80.6 & 68.2 & 73.9 & 87.6 & 79.9 & 83.6 & 86.0 & 81.1 & 83.5 & 88.2 & 78.2 & 82.9 \\

PSENet \cite{wang2019shape} & CVPR19 & \checkmark & & 75.3 & 69.2 & 72.2 & 84.0 & 78.0 & 80.9 & 84.8 & 79.7 & 82.2 & & & \\

FCE \cite{Zhu_2021_CVPR} & CVPR21 & \checkmark & & - & - & - & 89.3 & 82.5 & 85.8 & 87.6 & 83.4 & 85.5 & - & - & - \\

MOST \cite{he2021most} & CVPR21 & \checkmark & & \textbf{82.0} & 72.0 & 76.7 & - & - & - & - & - & - & 90.4 & 82.7 & 86.4 \\

ABPNet \cite{Zhang_2021_ICCV} & ICCV21 & \checkmark & & - & - & - & \textbf{90.67} & 85.19 & 87.85 & 87.66 & 80.57 & 83.97 & 86.62 & 84.54 & 85.57 \\

CentripetalText \cite{sheng2021centripetaltext} & NeurIPS21 & \checkmark & & - & - & - & 90.6 & 82.5 & 86.3 & \textbf{88.3} & 79.9 & 83.9 & 90.0 & 82.5 & 86.1 \\

PCR \cite{Dai_2021_CVPR} & CVPR21 & \checkmark & & & & & 88.5 & 82.0 & 85.2 & 87.2 & 82.3 & 84.7 & \textbf{90.8} & 83.5 & 87.0 \\ \hline

% our word
\multirow{2}{*}{Ours (word)} & \multirow{2}{*}{-} & \checkmark & & 77.71 & 75.88 & 76.78 & 85.49 & 90.53 & \textbf{87.94} & - & - & - & - & - & - \\

 & & \checkmark & \checkmark & 78.05 & \textbf{76.44} & \textbf{77.24} & 84.96 & \textbf{91.06} & 87.90 & - & - & - & - & - & - \\ \bottomrule
 
 % our line
 \multirow{2}{*}{Ours (line)} & \multirow{2}{*}{-} & \checkmark & & - & - & - & - & - & - & 83.92 & 85.87 & 84.88 & 86.59 & 86.81 & 86.69 \\
 
 & & \checkmark & \checkmark & - & - & - & - & - & - & 84.56 & \textbf{87.44} & \textbf{85.97} & 88.04 & \textbf{87.44} & \textbf{87.70} \\ \bottomrule
\end{tabular}}
\caption{Results of word and text line detection on public scene text datasets. Both our word and line detectors are outperforming the latest methods, even though our models are not fine-tuned for any target datasets. The proposed new dataset also proves to be a helpful complement to existing scene text datasets.}
\label{tab:detection-only}
%\vspace{3mm}
\end{table*}

In this section, we do ablation studies to further explore the design details. Except the detection granularity experiments (i.e. word v.s. line), we use $N=128$ object queries.

\noindent \textbf{Word-based v.s. line-based}: Our unified detector framework is able to perform end-to-end text entity detection on either word or line level, and then cluster these entities into the paragraph level as layout analysis results. 
Though word and line detections are largely interchangeable in terms of the subsequent recognition algorithms, we observe significant difference in layout analysis as shown in Tab. \ref{tab:word-line}. 
While both word and line level models benefit from more object queries, line level models consistently outperform their word level peers. 
One potential cause may be that detecting at line level reduces the number of objects compared to word-level detections, making the optimization for the clustering head easier.

\begin{table}[]
\scalebox{0.8}{\begin{tabular}{c|cclclc}
\hline
& \multicolumn{5}{c}{Text Line Detection} & \\
\multirow{-2}{*}{Text clustering head} & P & R & \multicolumn{1}{c}{F} & T & \multicolumn{1}{c}{PQ} & \multirow{-2}{*}{\begin{tabular}[c]{@{}c@{}}Layout \\(PQ)\end{tabular}} \\ \hline
Line detector only & \multicolumn{1}{l}{76.21} & \multicolumn{1}{l}{75.11} & 75.66 & \multicolumn{1}{l}{77.38} & 58.55 & - \\ \hline
no-extra layer & 75.50 & 74.16 & 74.82 & 77.14 & 57.72 & 48.07 \\
1 FC-ReLU-BN & 75.71 & 74.85 & 75.28 & 77.41 & 58.28 & {\color[HTML]{000000} 47.79} \\
MHSA x1 & 76.11 & 75.65 & 75.88 & 77.43 & 58.76 & {\color[HTML]{000000} 51.00} \\
MHSA x3 & 76.32 & 75.20 & \multicolumn{1}{c}{75.76} & 77.57 & \multicolumn{1}{c}{\textbf{58.76}}  & {\color[HTML]{000000} \textbf{51.48}} \\ \hline
\end{tabular}}
\caption{The impact of different text clustering head architecture on text and layout metrics.}
\label{tab:ablation-head}

% \vspace{1mm}

\end{table}

\noindent \textbf{Text clustering loss}: We compare the use of different ways to balance the clustering loss. Results are listed in Tab. \ref{tab:ablation-loss}. 
$\alpha$-balancing is the default method described in Sec. \ref{sec:loss}.
Vanilla means no balancing at all; it normalizes the loss term directly by $w = [\sum_{i=1}^N\sum_{j=1}^Ny_{\sigma(i)}y_{\sigma(j)}]^{-1}$. 
Applying $\alpha$-balancing factor achieves considerable improvements in both text detection and layout analysis.
Balancing the loss with focal style factors \cite{lin2017focal} results in worse performance in both tasks.

\noindent \textbf{Text clustering head}: We compare our default setting, a 3-layered multi-head self-attention (MHSA) \cite{vaswani2017attention} head, with other viable choices, as shown in Tab. \ref{tab:ablation-head}. We also list the performance of a MaX-DeepLab line detector without layout analysis branch. If we do not use any extra layer, the text detection performance is deteriorated compared to line detector only, indicating that it is necessary to separate the features. 
However, using fully connected layer cannot fully recover the ability to detect text and worsens layout analysis.
Using 1 layer of MHSA is better than only using fully-connected layer in both the detection and layout tasks. This is intuitive since Transformer's  \cite{vaswani2017attention} architecture block provides stronger modelling of interactions between text entities. 
Finally, additional transformer layers improve the performance.

\subsection{Scene text detection on public datasets}
In this section, we evaluate our models on the most widely used benchmarks for scene text detection. We adopt the same training and optimization settings in Sec. \ref{exp:unified} except that the layout analysis branch is excluded since other public datasets do not have layout labels. We use $N=384$ object queries. We do not initialize our models from any checkpoint. Nor do we pretrain on any synthetic datasets. We directly train on the union of public datasets without fine-tuning on any of them\footnote{For word detection, we use TextOCR\cite{Singh_2021_CVPR}, MLT17\cite{nayef2017icdar2017}, Total-Text\cite{ch2017total}, HierText. For line detection, we use LSVT\cite{sun2019icdar}, CTW1500\cite{yuliang2017detecting},
MSRA-TD500\cite{yao2012detecting}, HierText.}. We directly evaluate the models with the checkpoint of last training iteration. We evaluate on the following $4$ benchmarks: \textit{MLT 17}\cite{nayef2017icdar2017}, \textit{Total-Text}\cite{ch2017total}, \textit{CTW1500}\cite{yuliang2017detecting}, and \textit{MSRA-TD500}\cite{yao2012detecting}. Results and comparison with previous papers are summarized in Tab. \ref{tab:detection-only}. Overall, our detectors are characterized by higher recall and lower precision compared to state-of-the-art methods. Notably, even though curve text takes up a very small proportion in the training datasets, our model still excels at both curved text datasets, CTW1500 and Total-Text, showing case the adaptability of the proposed method.

For word detection, we achieve state-of-the-art result ($77.24$) on MLT 17. The performance is still very competitive ($76.78$) when trained on other public datasets only. On Total-text, we achieve state-of-the-art regardless of the use of HierText ($87.94$ and $87.90$). 

For line detection, we achieve very competitive results on CTW1500 and MSRA-TD500 without training on HierText. We observe considerable improvements when we add HierText in our training data ($84.88\rightarrow85.97$ and $86.69\rightarrow87.70$). This demonstrates that HierText is a helpful complement to the collection of public line datasets.

% ----- ----- ----- ----- ----- Conclusion ----- ----- ----- ----- ----- 
\section{Conclusion}
\label{sec:conclusion}
In this paper, we motivate the task of unified scene text detection and layout analysis. To facilitate research into this direction, we collect a dataset with hierarchical text annotations. We further propose an end-to-end model for unified detection and layout analysis that outperforms previous methods while at the same time greatly simplifying the pipeline. 
With the new task, dataset, and model, we push the envelop of text extraction and understanding in images and enable better support for downstream tasks.

%%%%%%%%% REFERENCES
{\small
\bibliographystyle{ieee_fullname}
\bibliography{egbib}
}

\end{document}